# Global optimization of expensive black-box models based on asynchronous hybrid-criterion with interval reduction

**Chunlin Gong[a]\*, Xu Li[a], Hua Su[a], Jinlei Guo[b], Liangxian Gu[a]**

\*Corresponding author: leonwood@nwpu.edu.cn

[a]School of Astronautics, Northwestern Polytechnical University, 710072, Xi'an, P.R. China

[b]Shanghai Electro-Mechanical Engineering Institute, 201109, Shanghai, P.R. China

**Abstract**: In this paper, a new sequential surrogate-based optimization (SSBO) algorithm is developed, which aims to improve the global search ability and local search efficiency for the global optimization of expensive black-box models. The proposed method involves three basic sub-criteria to infill new samples asynchronously to balance the global exploration and local exploitation. First, to capture the promising possible global optimal region, searching for the global optimum with genetic algorithm (GA) based on the current surrogate models of the objective and constraint functions. Second, to infill samples in the region with sparse samples to improve the global accuracy of the surrogate models, a grid searching with Latin hypercube sampling (LHS) with the current surrogate model is adopted to explore the sample space. Third, to accelerate the local searching efficiency, searching for a local optimum with sequential quadratic programming (SQP) based on the local surrogate models in the reduced interval, which involves some samples near the current optimum. When the new sample is too close to the existing ones, the new sample should be abandoned, due to the poor additional information. According to the three sub-criteria, the new samples are placed in the regions which have not been fully explored and includes the possible global optimum point. When a possible global optimum point is found, the local searching sub-criterion captures the local optimum around it rapidly. Numerical and engineering examples are used to verify the efficiency of the proposed method. The statistical results show that the proposed method has good global searching ability and efficiency.

## 1 Introduction

With the development of computing technology, it is possible to simulate the behavior of complex systems with high-fidelity models. However, when it comes to finite element analysis (FEA) and computational fluid dynamics (CFD), in which a single simulation may cost several hours, the optimization process is challenging. Additionally, the high-fidelity model may be from expensive experiments, such as the large wind tunnel experiment. Therefore, the high-fidelity model is time cost and economic expensive. On the other hand, the traditional gradient-based optimization algorithm, such as sequential quadratic programming (SQP) is fast but may lose the global optimum, while the global algorithm, such as genetic algorithm (GA), is too expensive, due to the thousands of high-fidelity model evaluations. In order to balance the computational efficiency and global searching ability in the optimization process of expensive black-box problems, the surrogate-based optimization (SBO) algorithm has gained significant attention recently (Floudas and Gounaris 2009; Koziel and Leifsson 2013; Vu et al. 2016). This kind of method can make a reasonable trade-off between the accuracy and efficiency with less cost under a certain strategy of adding samples, which has been discussed extensively in the literature.

As one of the earliest scholars who studied SBO, Jones et al. (1998) proposed an efficient global optimization (EGO) algorithm that maximizes the expected improvement (EI) function of Kriging to add samples. The next new sample is selected where the expectation of the largest improvement is located. EGO works well to solve the expensive black-box low-dimensional unconstrained optimization problem. Huang (2006) proposed the sequential Kriging optimization (SKO) method which is an extension of the EGO. SKO modifies the optimization function of adding samples. It was reported that SKO is remarkable when the objective function is smooth but may be degraded to a local search algorithm when it comes to multi-peak problems. Booker et al. (1999) developed a general framework named surrogate management framework (SMF) based on the idea of SBO, and the convergence of SMF was proved. Gutmann et al. (2001) presented a radial basis function (RBF) based global optimization method for expensive black-box non-convex continuous functions, adding samples at the current optimum of the surrogate model. Regis and Shoemaker (2007) used RBF to construct a new and efficient evaluation function which limited the region of adding samples. Holmström (2008) proposed a new adaptive radial basis function (ARBF) method and developed a toolkit containing several commonly used algorithms. Moreover, Stefan et al. (Jakobsson et al. 2010) developed similar modified methods. Kieslich et al. (2018) presented a method based on sparse grids and polynomial approximations, which aims to locate the global optimum of box-constrained problems using input-output data. Jones (2001) compared several different SBO methods and discussed the global searching ability and failure situations. Robinson et al. (2008) applied a multi-fidelity surrogate model to the engineering optimization problem, the optimization time of which was reduced a lot. Haftka (2016) surveyed the parallel application of SBO and gave some suggestions to choose an appropriate method when it came to engineering applications. Li et al. 2016) investigated the parallel algorithms in efficient global optimization based on multiple points infill criterion and domain decomposition. The method can reduce the overall computing time, but it consumes more computing resources and is difficult to apply when the computing resources are limited. Liu (2018) made the latest review of SBO and discussed the factors to affect the performance of SBO. In addition to the methods mentioned above, more criteria of adding points, such as surrogate lower bound (SLB), maximum probability (MP), surface minimization (SURF$_{min}$), candidate point approach (CPA) et al., can be found in the literatures (Booker et al. 1999; Floudas and Gounaris 2009; Haftka et al. 2016; Koziel and Leifsson 2013; Liu 2017; Liu et al. 2016; Mueller 2014; Sun et al. 2015; Vu et al. 2016).

SBO algorithms use surrogate models instead of the original expensive black-box models for optimization. They add samples to refine the local regions where global optimum may exist. The key issue of SBO is how to accelerate the algorithm convergence speed and ensure the global search ability and adaptability to different problems. Most traditional sequential sampling methods construct a special function, and by optimizing the function automatically find points that can weigh the local and global search capabilities. The construction of this function is usually an optimization problem with strong constraints on the multimodal function, and it is difficult to obtain the theoretical optimum when solving the problem. While other sequential methods only add points near the current best point, the methods often lead to premature convergence of the algorithm in some local areas and are highly dependent on the early iteration process. This problem is very prominent in strong nonlinear problems.

In this paper, a hybrid criterion is developed to infill samples sequentially to solve the global optimization problem with black-box constraint functions. In the proposed criterion, three sub-criteria are involved. First, searching for the global optimum with GA based the current surrogate model. Second, searching for local optimum with SQP based on the current surrogate model. Here, the searching space is a neighborhood of the current optimum in the existing samples. The neighborhood is a region including a constant number of samples. Third, to exploit the sample space with a strict minimum distance constant, a grid searching with Latin hypercube sampling (LHS) with the current surrogate model is adopted. This sub-criterion aims to infill samples in the region with sparse samples.

The remainder of this paper is structured as follows. In Section 2, the surrogate model used in the proposed method is introduced. In Section 3, the details of the proposed method are presented and discussed. In Section 4, several examples including mathematical and engineering ones are used to demonstrate the efficiency, accuracy of the proposed method. Meanwhile, the results are compared with the existing typical SBO methods. Finally, Section 5 discusses remarks and further research directions.

## 2 Surrogate model: Radial basis function

2.1 Build the RBF model

A surrogate model is an approximated prediction model based on the training samples. Instinctually, it is an interpolation or regression model, which is a branch of machine learning (Hastie and Tibshirani 2008). As a typical surrogate model, RBF has good nonlinear adaptability. Moreover, RBF has only one hyper-parameter, which illustrates RBF is easier to construct. Assume that the observation samples are presented as $S = \{(\mathbf{x}_i, y_i) | \mathbf{x}_i \in \mathbb{R}^m, y_i \in \mathbb{R}^1, i = 1,2,\cdots,n\}$, where $\mathbf{x}_i$ is the input variable, $y_i$ is the response variable, $n$ is the sample size, and $m$ is the dimension of the input variable. RBF uses a series of linear combinations of radial basis functions to approximate the expensive black-box function, which can be formulated as

$$\hat{y}(\mathbf{x}) = \sum_{i=1}^{n} \beta_i \, \mathrm{f}\left(\|\mathbf{x}-\mathbf{x}_i\| | c\right) = \mathbf{f}(\mathbf{x} | c)^{\mathrm{T}} \boldsymbol{\beta} \tag{1}$$

Table 1 Radial basis functions.

| Type | Function form $f(r)$ |
| --- | --- |
| Gaussian | $\exp(-cr^2)$ |
| Multi-quadric | $(1+cr^2)^{1/2}$ |
| Inverse Multi-quadric | $(1+cr^2)^{-1/2}$ |
| Thin plate spline | $r^2\log(1+cr^2)$ |

where $\hat{y}(\mathbf{x})$ denotes the predictive response at point $\mathbf{x}$, $\beta_i$ is the $i$th component of the radial basis coefficient vector $\boldsymbol{\beta}$, and $f(\|\mathbf{x} - \mathbf{x}_i\| | c)$ (See Table 1) is the $i$th component of the radial basis function vector $\mathbf{f}(\mathbf{x}|c)$. $r = \|\mathbf{x} - \mathbf{x}_i\|$ is the Euclidian distance between two samples, and $c$ is the hyper-parameter. Substitute the samples into Eq. (1),

$$\begin{bmatrix} y_1 \\ y_2 \\ \vdots \\ y_n \end{bmatrix} = \begin{bmatrix} f(r_{11}) & f(r_{12}) & \cdots & f(r_{1n}) \\ f(r_{21}) & f(r_{22}) & \cdots & f(r_{2n}) \\ \vdots & \vdots & \ddots & \vdots \\ f(r_{n1}) & f(r_{n2}) & \cdots & f(r_{nn}) \end{bmatrix} \begin{bmatrix} \beta_1 \\ \beta_2 \\ \vdots \\ \beta_n \end{bmatrix} \qquad (2)$$

Rewritten as a matrix form

$$\mathbf{y} = \mathbf{F}\boldsymbol{\beta} \qquad (3)$$

where $\mathbf{y} = [y_1, y_2, \cdots, y_n]^T$, $\mathbf{F} = [f(r_{ij})]_{n \times n}$. As the samples are different from each other, $\mathbf{F} \in \mathbb{R}^{n \times n}$ is a non-singular matrix, therefore Eq. (2) has a unique solution $\boldsymbol{\beta} = \mathbf{F}^{-1}\mathbf{y}$. Thus, the prediction model is given by

$$\hat{y}(\mathbf{x}) = \mathbf{f}(\mathbf{x}|c)^T \mathbf{F}^{-1} \mathbf{y} \qquad (4)$$

where $\mathbf{f}(\mathbf{x}|c)$ is related to the prediction point $\mathbf{x}$ and sample input matrix $\mathbf{X} = [\mathbf{x}_1, \mathbf{x}_2, \cdots, \mathbf{x}_n]^T$, $\mathbf{F}^{-1}\mathbf{y}$ is only related to $\mathbf{X}$ and $\mathbf{y}$. For a new prediction sample $\mathbf{x}$, $\mathbf{f}(\mathbf{x}|c)$ is calculated one time to get its predicted value $\hat{y}(\mathbf{x})$. It should be pointed out that the hyper-parameter $c$, which has a great influence on the accuracy of the model, is included in $\mathbf{F}$ and is determined by experience or other optimization criteria. Since this paper deals with expensive black-box problems, the number of the training samples is small, which illustrates that there are no extra samples for model validation. Therefore, the hyper-parameter $c$ can be obtained by optimizing the cross-validation error, which does not require additional validation samples other than the training samples.

2.2 Optimize the hyper-parameter

To avoid extra samples, the cross-validation (CV) method is adopted to validate the surrogate model (Hastie and Toshigami 2008; Viana et al. 2010). The samples are divided into $K$ roughly equal-sized parts. For the $k$th part ($k$ = 1, 2, $\cdots$, $K$), the other $K$−1 parts of the samples are used to construct the surrogate models, and the $k$th part samples are used to obtain the $k$th prediction error. When $K$ = $n$, it is called leave-one-out cross-validation error (LOOCV). The hyper-parameter $c$ in Eq. (4) can be estimated by the following optimization problem:

$$\min_c \sum_{i=1}^{n} \left[ y_i - \hat{y}\left(\mathbf{x}_i | S - \{(\mathbf{x}_i, y_i)\}, c\right) \right]^2 \qquad (5)$$

In Eq. (5), the evaluation of LOOCV error requires $n$ times construction of the surrogate model. However, the LOOCV error does not require additional verification points, which is capable of describing the matching degree between the samples and the prediction model. According to the surrogate model and obtained parameter $c$, an optimization method based on the sequential surrogate model can be developed.

# 3 The proposed method

In practical applications, it is difficult to validate whether the current optimum is the global of local optimum or not. If the current optimum is also the global one, the convergence accuracy and speed will be good. However, if the current optimum is not the global optimum, it will not help to refine the local region. In this article, in order to reduce the computational difficulties to search for the pseudo-global optimum, a hybrid-criterion method of adding samples is proposed. Different optimization algorithms are adopted to search the optimum based on the criterions. The proposed method aims to improve the local accuracy while searching for the global optimum.

3.1 Hybrid-criterion for exploitation and exploration

3.1.1 Minimization criterion for global optimization (**Criterion 1**)

As surrogate models capture some information of the original expensive black-box function, the global optimum of the surrogate models also captures the information of the real global optimum. With the iteration goes on, the approximation of the global optimum gets closer to the real solution. This process is formulated as:

$$\mathbf{x}_k = \operatorname*{argmin}_{\mathbf{x}} \hat{J}(\mathbf{x}|S_{k\text{-}1})$$
$$\text{s.t.} \begin{cases} \hat{\mathbf{g}}(\mathbf{x}|S_{k\text{-}1}) \leq \mathbf{0} \\ \mathbf{x}_L \leq \mathbf{x} \leq \mathbf{x}_U \end{cases} \tag{6}$$

where $\mathbf{x} \in \mathbb{R}^n$, $\mathbf{x}_k$ denotes the sample added in the $k$th iteration, $\hat{J}(\mathbf{x}|S_{k-1})$ is the surrogate objective function based on the sample set $S_{k-1}$, $\hat{\mathbf{g}}(\mathbf{x}|S_{k-1})$ is the surrogate constraint vector function based on $S_{k-1}$, $\mathbf{x}_L$ and $\mathbf{x}_U$ are the lower and upper bounds of $\mathbf{x}$, respectively.

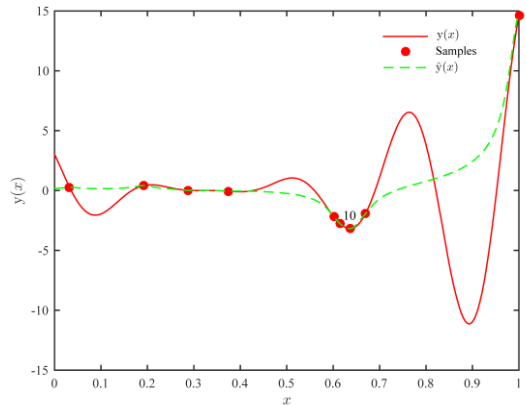

Fig. 1 The process of convergence to a local optimum.

Equation (6) aims to search for the global optimum based on the surrogate models with global optimization algorithm such as the genetic algorithm (GA). However, this criterion depends on the accuracy of the initial surrogate models. If the initial surrogate models cannot capture the approximation region of the real global optimum, this criterion converges to a local optimum. As shown in Fig. 1, the criterion converges to a local optimum after ten iterations, due to the poor global accuracy of the surrogate models. More criterions should be adopted to improve the global accuracy of the surrogate models, so that the surrogate models have the ability to capture the region information of the real global optimum.

3.1.2 Uniform criterion for global accuracy (**Criterion 2**)

The uniform criterion aims to improve the global accuracy of the surrogate models. It adds samples where the surrogate models have the poorest accuracy. According to Tayler's series expansion, the prediction error increases with the distance between the prediction sample and the existing samples. The distance between a prediction point **x** and the existing samples is defined as:

$$\mathrm{d}(\mathbf{x} \mid S_{k-1}) = \min \left\{ \|\mathbf{x} - \mathbf{x}_i\| \mid (\mathbf{x}_i, \mathbf{J}(\mathbf{x}_i), \mathbf{g}(\mathbf{x}_i)) \in S_{k-1} \right\} \tag{7}$$

Therefore, the point, which is furthest from the existing samples, is selected to improve the global accuracy of the surrogate models (Fig. 2). The criterion is formulated as

$$\mathbf{x}_k = \underset{\mathbf{x}}{\operatorname{argmin}} \left[ \mathrm{d}(\mathbf{x}|S_{k-1}) \right]^{-1}$$
$$\mathrm{s.t.}\, \mathbf{x}_L \leq \mathbf{x} \leq \mathbf{x}_U \tag{8}$$

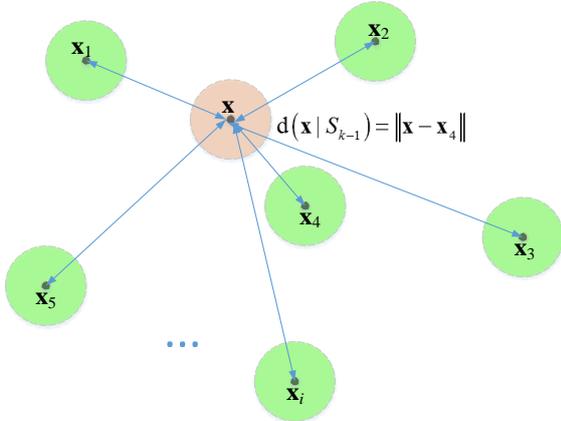
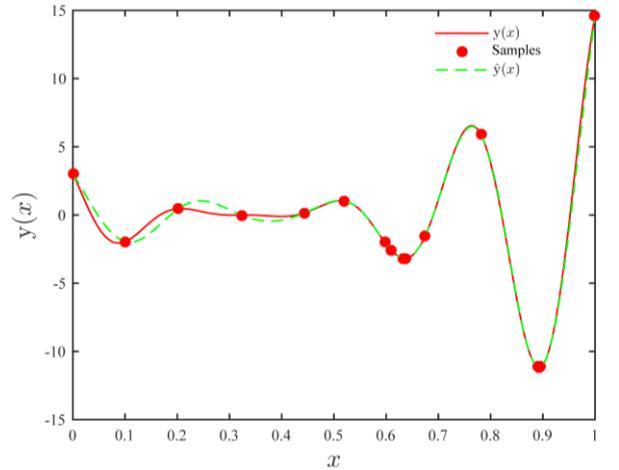

Fig. 2 The geometric meaning of the minimum distance.

Fig. 3 Uniform criterion which captures the global optimal region.

As shown in Fig. 3, since the objective function of this criterion is a multi-peak function with many local optimal solutions, it is difficult for conventional gradient optimization algorithm to obtain the global optimum, and sometimes it is even difficult to obtain the feasible solution in constraint optimization problem. Although the genetic algorithm can find the global optimal solution in theory, its performance is not satisfactory in solving such problems here. Since it is not necessary to obtain a very accurate global optimal solution, the main purpose of the algorithm is to obtain a relatively uniform sample distribution, to improve the global accuracy of the surrogate models.

Therefore, the direct sampling method is adopted here as an approximated optimization criterion. Firstly, a sample set is generated by the Latin hypercube sampling (LHS) method, and then the grid search method is adopted to find the optimum. The optimization problem is transformed into:

$$\mathbf{x}_k = \underset{\mathbf{x}_i \in D}{\operatorname{argmin}} \left[ d(\mathbf{x}_i | S_{k\text{-}1}) \right]^{-1}$$
$$\text{s.t.} \begin{cases} D = \{\mathbf{x}_1, \mathbf{x}_2, \cdots, \mathbf{x}_{N_{\text{LHS}}}\} \\ N_{\text{LHS}} = 1 \times 10^4 \end{cases} \quad (9)$$

where $D$ is the sample set of LHS, $N_{\text{LHS}}$ is the number of the simulations. This criterion does not depend on the surrogate models, which keeps the tendency of uniform. As shown in Fig. 3, with the cooperative performance of uniform criterion, the algorithm captures the region which includes the real global optimum. This criterion ensures the algorithm explores the global region when the minimization criterion does not find a better point, which avoids being trapped in a local optimum.

### 3.1.3 Interval reduction criterion for accelerating local convergence (**Criterion 3**)

When the surrogate model is complex, the global optimization algorithm cannot increase the local search ability well, and the obtained optimization solution fluctuates greatly in the local region, making the convergence slow in the local search. Therefore, gradient-based optimization algorithm with high accuracy of local search is adopted in a certain region. Sequential quadratic programming can solve the following optimization problems:

$$\mathbf{x}_k = \underset{\mathbf{x}}{\operatorname{argmin}} \hat{J}(\mathbf{x}|T_{k\text{-}1})$$
$$\text{s.t.} \begin{cases} \hat{\mathbf{g}}(\mathbf{x}|T_{k\text{-}1}) \leq \mathbf{0} \\ \mathbf{x}_L^{T_{k\text{-}1}} \leq \mathbf{x} \leq \mathbf{x}_U^{T_{k\text{-}1}} \end{cases} \quad (10)$$

where $\mathbf{x}_k$ is the point added in the $k$th iteration, $\mathbf{x}_L^{T_{k-1}}$ and $\mathbf{x}_U^{T_{k-1}}$ are lower and upper bounds for $\mathbf{x}$. $T_{k\text{-}1}$ is the important sample set, which is a subset of the current sample set $S_{k\text{-}1}$. $T_{k\text{-}1}$ includes the closest $p$ samples to the current optimum $\mathbf{x}_{k-1}^*$. The larger $p$ is, the larger the important region is, the slower the algorithm converges, and the less likely it is to fall into the local optimum. The smaller $p$ is, the faster the algorithm converges, but it is easy to cause premature convergence. In this paper, $p$ is selected to be equal to the number of the initial sample set. The formulation of the importance region is given as:

$$T_{k-1} = \left\{ \left( \mathbf{x}_{(i)}, J(\mathbf{x}_{(i)}), \mathbf{g}(\mathbf{x}_{(i)}) \right) | 1 \leq i \leq p \right\} \quad (11)$$

where $\mathbf{x}_{(i)}$ is the reordered sample according to the distances to the current optimum $\mathbf{x}_{k-1}^*$ from small to large. The arrangement of reordered distances is:

$$\left\| \mathbf{x}_{(1)} - \mathbf{x}_{k-1}^* \right\| < \left\| \mathbf{x}_{(2)} - \mathbf{x}_{k-1}^* \right\| < \cdots < \left\| \mathbf{x}_{(p)} - \mathbf{x}_{k-1}^* \right\| < \cdots < \left\| \mathbf{x}_{(n)} - \mathbf{x}_{k-1}^* \right\| \quad (12)$$

This local criterion takes a part of the samples to build the surrogate models and search the local optimum with the gradient-based optimization algorithm within a reduced interval. A local surrogate model has better local accuracy, and the gradient-based algorithm has better local accuracy, which accelerates the local convergence of the criterion. As shown in Fig. 4, though this criterion accelerates the local search process, it may be trapped in a local optimum. The uniform criterion should also work together with the uniform criterion for global accuracy.

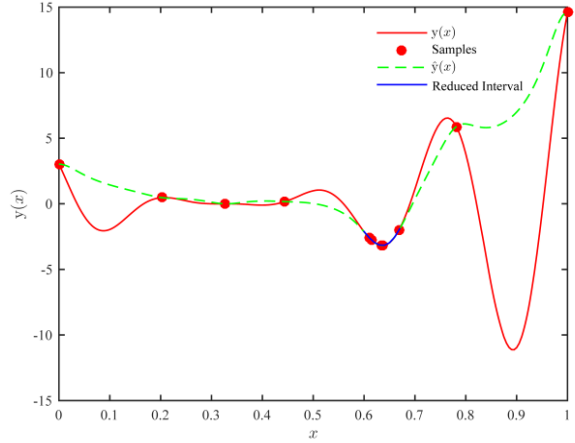

Fig. 4 Surrogate models in the reduced interval.

Sometimes **Criterion** 1 and **Criterion** 3 obtain samples, which are too close to the existing samples to supply extra information to refine the surrogate models. To reduce the unnecessary calculation, the current point should be abandoned when $d(\mathbf{x}|S_{k-1}) < \varepsilon (= 1 \times 10^{-3})$. Therefore, there is no need to take the expensive function evaluation and update the sample set $S_{k-1}$.

3.2 The procedure of the proposed method

This section proposes a sequential surrogate-based optimization (SSBO) algorithm. The core idea of SSBO is to use the hybrid-criterion to enhance the global search and use an interval reduction method to speed up the local search. As shown in. Fig 5, the main step of SSBO is an iteration process of adding samples. The more detailed steps are shown as follows:

**Step 1:** Select the initial input samples $\mathbf{x}_i (i = 1, 2, \cdots, n_0)$ with the design of experiment (DoE) method Latin hypercube sampling (LHS). There are two reasons for choosing LHS. First, LHS can obtain well-distributed and representative design points with less cost from the design space, so as to obtain accurate model information more effectively. Second, LHS has the freedom to define the number of sample points, thereby providing greater flexibility and broader applicability. The number of the initial samples is chosen as $n_0 = 2m + 1$, where $m$ is the dimension of the design variables. Evaluate the input samples with the expensive black-box model to obtain the response values of the objective J($\mathbf{x}_i$) and constraint vector **g**($\mathbf{x}_i$). Thus, the initial sample set $S_0$ is given by

$$S_0 = \{(\mathbf{x}_i, J(\mathbf{x}_i), \mathbf{g}(\mathbf{x}_i)) | i = 1, 2, \cdots, n_0\} \tag{13}$$

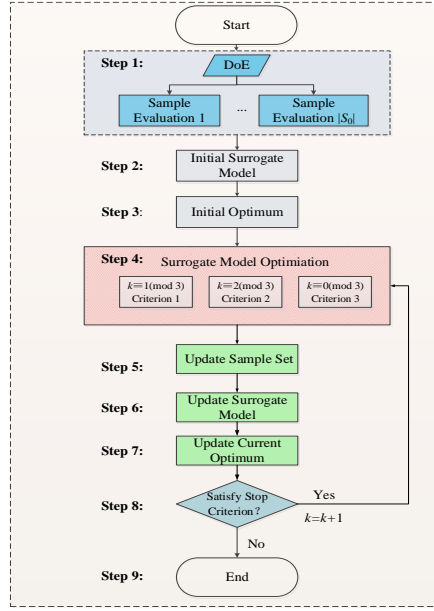

Fig 5. The procedure of the sequential surrogate-based optimization method.

**Step 2:** Use the initial sample set $S_0$ to construct the surrogate objective function $\hat{J}(\mathbf{x}|S_0)$ and the surrogate constraint vector $\hat{\mathbf{g}}(\mathbf{x}|S_0)$. Then the shape parameters of the surrogate models are optimized with cross-validation. Here the surrogate models are constructed with RBF, which strictly goes through the samples and has strong nonlinear adaptability. The RBF code is complemented with the authors' in-house MATLAB toolbox.

**Step 3:** Search for the initial optimum. Select the optimum which satisfies the constraints in the sample set $S_0$. If none of the points in the sample set satisfies the constraints, then select the point with minimum objective function value as the initial optimum.

**Step 4:** Search for the new point based on the hybrid-criterion of adding points. Since the time cost of a surrogate model is much less than that of the real expensive black-box model, different optimization algorithm methods can be adopted for different criterions. For more details of the hybrid-criterion of adding points, refer to Section 3.1.

**Step 5**: Evaluate the real objective and constraint functions with the new point found in Step 4, and then update the sample set:

$$S_k = S_{k-1} \cup \left\{\left(\mathbf{x}_k, \mathbf{J}(\mathbf{x}_k), \mathbf{g}(\mathbf{x}_k)\right)\right\} \tag{14}$$

**Step 6:** Update the objective and constraint RBF surrogate models with the new sample set in Step 5, and the procedure is similar with that of Step 2.

**Step 7:** Update the current optimum. Search for the optimum that satisfies the original constraints with the exhaustive method. The sample newly found is taken as the current optimum.

**Step 8**: Stop criterion check. If the number of the added points $k$ reaches the required maximum number $k_{\max}$, go to Step 7, otherwise, return to Step 4.

**Step 9:** End of the algorithm.

3.3 Discussion of the algorithm

The core steps of the algorithm are the choice of surrogate models and the criterion of adding points. Surrogate models with high interpolation accuracy can capture the region containing the global optimum information according to the added sample points. If the regression method is used, the information added by the single sample points may be filtered by the algorithm. The selection of the criterion considers the global accuracy of the surrogate model and the local search accuracy. The difference from the existing criterion of adding points is that SSBO does not balance the global search and the local search through a single function, but directly adopts different criterions asynchronously to add points. This approach does not require additional computational resources for parallel computation and can effectively guarantee the global convergence and local convergence accuracy of the algorithm. At the same time, in order to improve the accuracy of local convergence, an important region (reduced interval) is determined near the current global optimum, and then a local surrogate model is constructed in the important region and a local optimization algorithm is used for the more accurate global optimum.

The main computation of the algorithm is the number of sample evaluations. If the time of a single evaluation of the surrogate model is much lower than the time of the expensive black-box function model, the total time required for the calculation is the evaluation time of the initial and added samples. The time to build the RBF surrogate model and to search for the new samples is negligible.

In terms of the convergence of the algorithm, criterion 2 of adding point search for the point in the sparse place, when the sample point is sufficiently dense, the RBF method for constructing the surrogate model can theoretically approximate an arbitrary continuous function (Buhmann 2009; Regis and Shoemaker 2005). Moreover, the genetic algorithm used in criterion 1 of adding point has a global convergence property (Liepins 1992). Therefore, the global convergence of the proposed method is based on the convergence of the RBF surrogate model and the genetic algorithm.

The main disadvantage of the algorithm is that the convergence precision cannot be controlled effectively, because the current global optimum is used as the criterion for convergence. For a higher degree of nonlinearity or a multi-peak problem, the current global optimum only changes after many iterations, so the maximum iteration number $k_{max}$ is adopted to terminate the algorithm. Theoretically, the larger $k_{max}$ is, the more possible to obtain the real global optimal solution. In actual problems, the number of calculations can be determined according to the computing resources and calculation time.

4 **Numerical examples**

In this section, several examples are used to perform SSBO, and the results are compared with those of some other existing methods.

## 4.1 1D mathematical problem

This example (Forrester et al. 2008) is a demonstration to validate the global searching performance of the proposed method. Mathematical functions of different complexities are considered. The formulation of the function is given by

$$y(x) = (6x - 2)^2 \sin(12x - 4) \cos(\alpha(x - 1)^2) \tag{15}$$

where $\alpha$ is a coefficient to describe the complexity of the function, and the larger $w$ illustrates the more complex function which obtains more local optimums. The function curve and iteration of response are shown in Fig. 6. As expected, with the increasing of $\alpha$, the number of iterations also increases. The detailed results are shown in Table 1. The function evaluation of SQP is the small least when $\alpha$ is small.

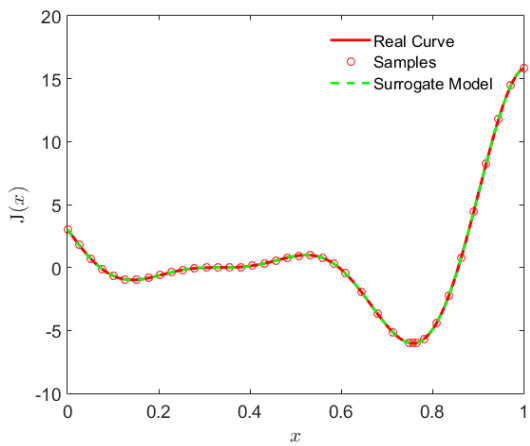

(a) Function curve and samples ($\alpha$=0).

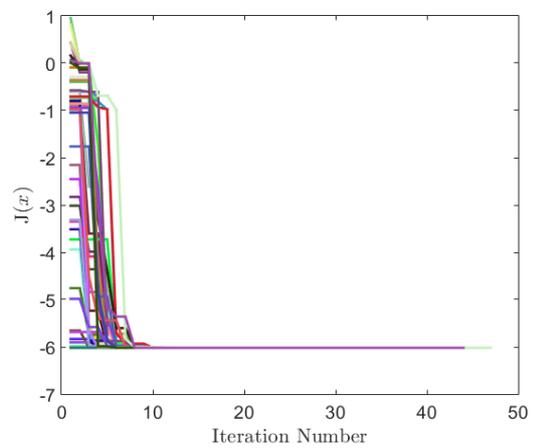

(b) Iteration of response ($\alpha$=0).

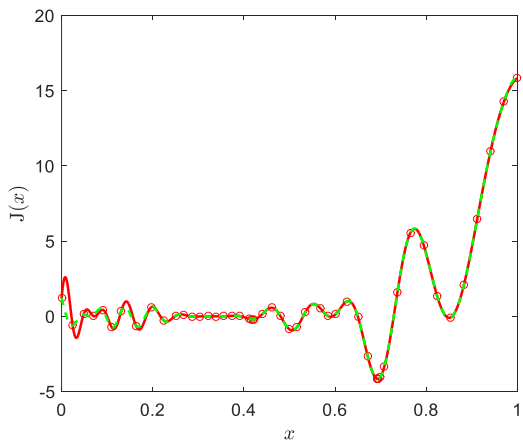

(c) Function curve and samples ($\alpha$=64).

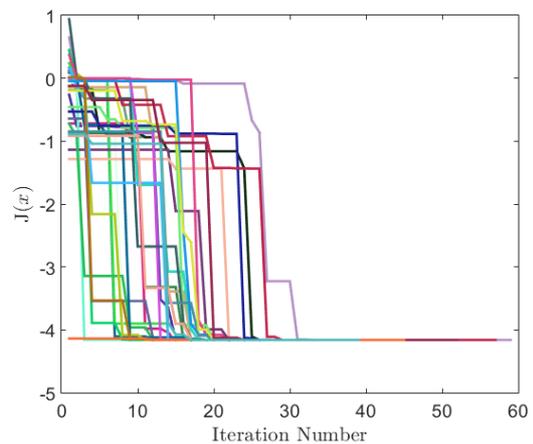

(d) Iteration of response ($\alpha$=64).

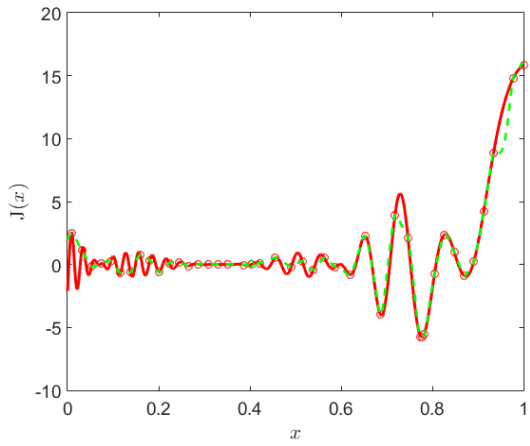

(e) Function curve and samples ($\alpha$=128).

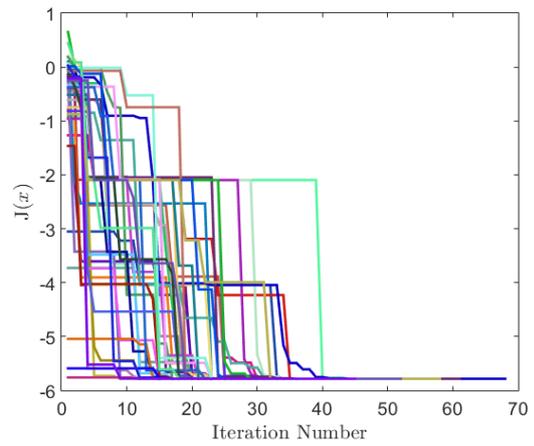

(f) Iteration of response ($\alpha$=128).

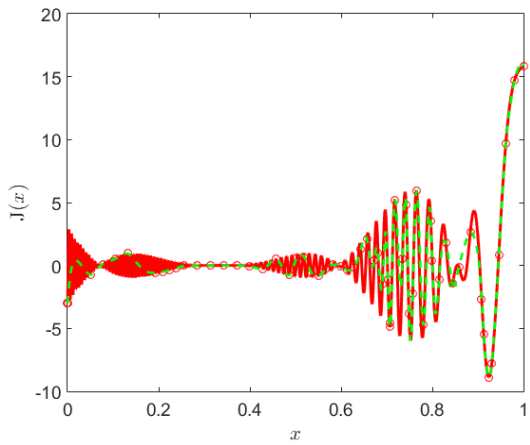

(g) Function curve and samples ($\alpha$=512).

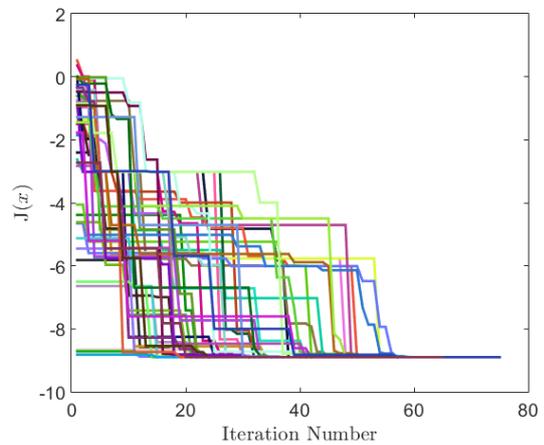

(h) Iteration of response ($\alpha$=512).

Fig. 6 Iterations for the 1D mathematical problem.

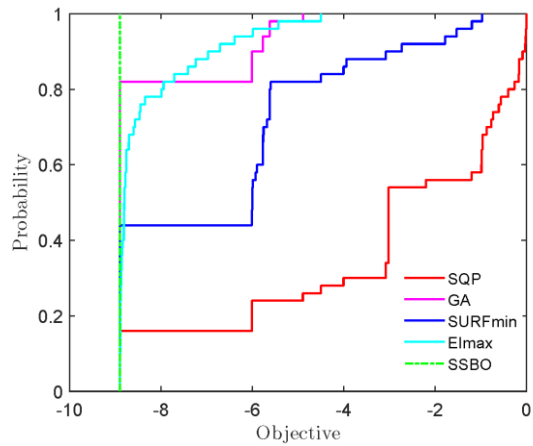

Fig. 7 The cumulative density functions for the 1D mathematical problem. ($\alpha = 512$)

Table 2 Comparison of different methods for the 1D mathematical problem. ($\alpha = 512$)

| Methods | Mean Value | Std. | Global Probability | $N$ |
| --- | --- | --- | --- | --- |
| SQP | -3.1607 | 3.0722 | 16% | 28.5 |
| GA | -8.3317 | 1.2324 | 82% | 3450 |
| SURF$_{min}$ | -6.5979 | 2.4073 | 44% | 103 |
| EI$_{max}$ | -8.3456 | 0.9952 | 22% | 103 |
| SSBO | -8.8988 | $4.1299 \times 10^{-6}$ | 100% | 61.2 |

Fig. 6 shows the model curve and the convergence curve of SSBO under different complexities. It can be seen that at the beginning of the iteration, due to the small number of initial samples and the low accuracy of the surrogate model, the improvement of the global surrogate model by increasing sample points is significant. With the increase of sample points, the change of the global optimum tends to converge. If the exploration during the iteration does not find a better point, the current optimum is not updated, but the surrogate model is updated. Therefore, the iteration curve remains unchanged for a period of time, and then the objective function value becomes smaller rapidly due to the optimal region captured by a certain iteration, showing the feature of jump change. When the initial samples are good, the algorithm converges to the global optimal solution quickly. When the initial samples are poor, the algorithm can also find the global optimum through certain global exploration points. Meanwhile, it can be seen that when $\alpha$ is greater, the model is more complex with more local optimums, and more samples are required for convergence.

Fig. 7 shows the statistical cumulative distribution function (CDF) of different algorithms when $\alpha = 512$. As shown in Fig. 7, the global optimal solution determined by random sampling is scattered around different local optimal solutions. The ideal case is a line perpendicular to the horizontal coordinate where the global optimum is, indicating that the probability of convergence to the global optimum is 100%. Actually, to obtain such a result, a large number of samples need to be calculated. After the samples reach a certain number, the probability of convergence does not increase significantly. At the same time, it can be seen that the CDF corresponding to different algorithms is different. When the increase speed is faster in the figure, the convergence is relatively more concentrated, and the global search ability is stronger.

As shown in Table 2, SQP has the smallest number of function evaluations, but since this problem has many local optimums, the objective function J(**x**) has a large mean value and standard deviation, and the global convergence probability is only 16%. GA has better objective value, but the standard deviation is large, this is because GA is based on a certain probability convergence to the global optimum. However, in the more complicated problem with many local optimums, GA is not easy to converge to the global optimum, and the global convergence probability is only 82%. SURF$_{min}$ is a criterion, which does not consider the global exploration. Although it converges quickly, due to different initial points, the probability of global convergence is only 44%. EI$_{max}$ is a criterion combining local and global search. Due to the multi-peak characteristic of the problem, the mean value of the objective function is small, but the standard deviation is large, indicating that the local accuracy is not high. Since the SSBO method proposed in this paper adopts global and local search algorithm rotation and flexible use of multiple optimization algorithms, it can ensure stable convergence to the global optimum within a given number of function evaluations. Therefore, the optimal standard deviation of the objective function is only 4.1299×10$^{-6}$, and the global convergence probability is 100%. Therefore, the global search optimization capability of SSBO has a significant advantage compared with the existing methods.

4.2 2D mathematical problem

This example (Cho and Lee 2011; Yi et al. 2016) is used to demonstrate the performance of SSBO algorithm in low dimensional constrained optimization problems. The example is a deterministic optimization problem with two optimization variables and three inequality constraints. The problem is described as follows:

$$\min_{\mathbf{x}} J(\mathbf{x}) = x_1 + x_2$$

$$\text{s.t.} \begin{cases} g_1(\mathbf{x}) = 1 - \dfrac{x_1^2 x_2}{20} \\ g_2(\mathbf{x}) = 1 - \dfrac{(x_1 + x_2 - 5)^2}{30} - \dfrac{(x_1 - x_2 - 12)^2}{120} \\ g_3(\mathbf{x}) = 1 - \dfrac{80}{x_1^2 + 8x_2 + 5} \\ 0 \le x_1, x_2 \le 10 \end{cases} \quad (16)$$

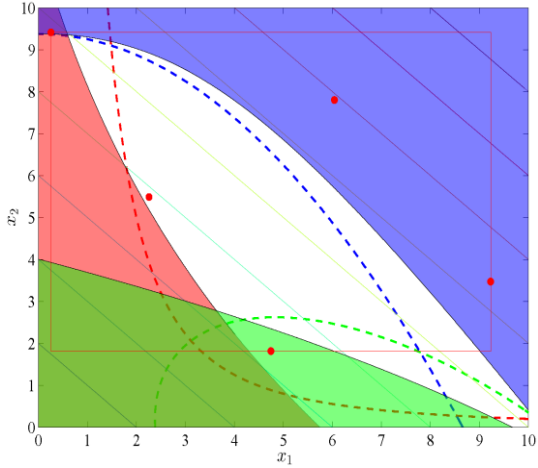 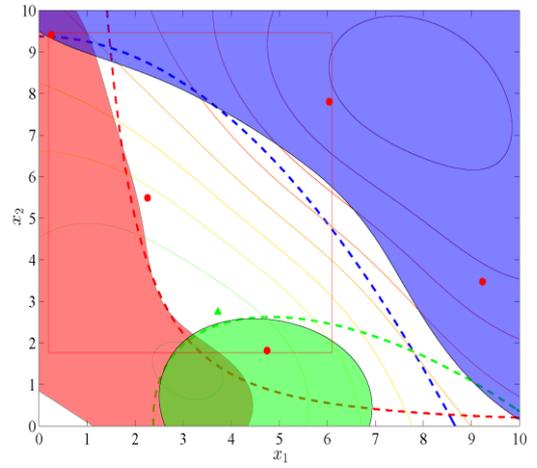

(a) initial state        (b) iteration #1

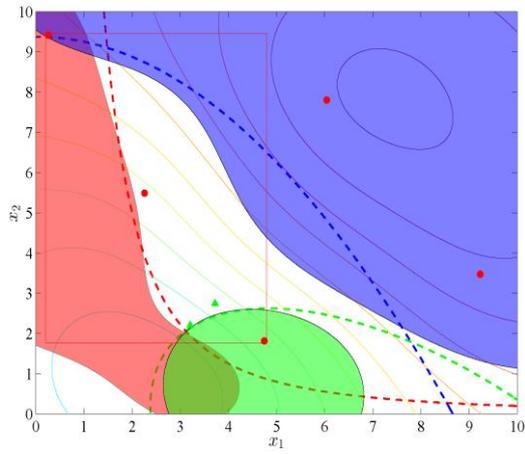

(c) iteration #2

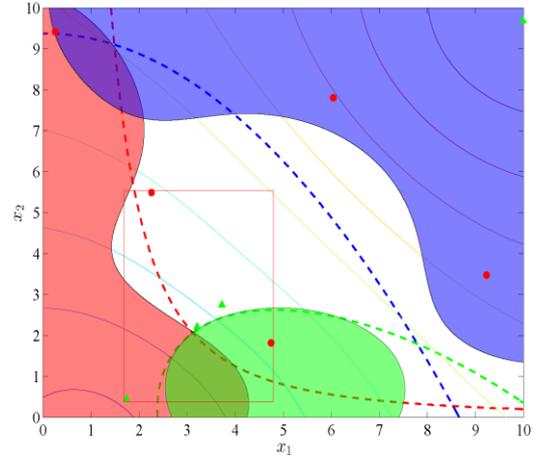

(d) iteration #3~4

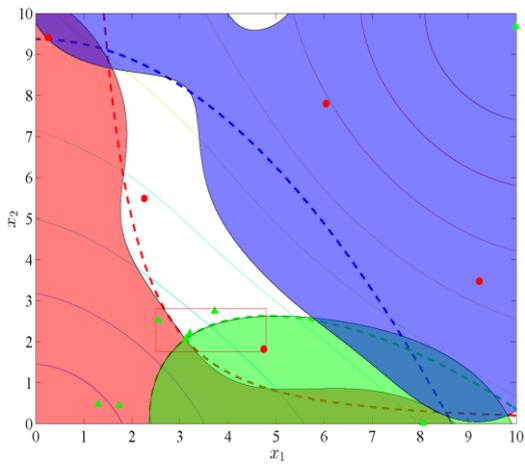

(e) iteration #5~8

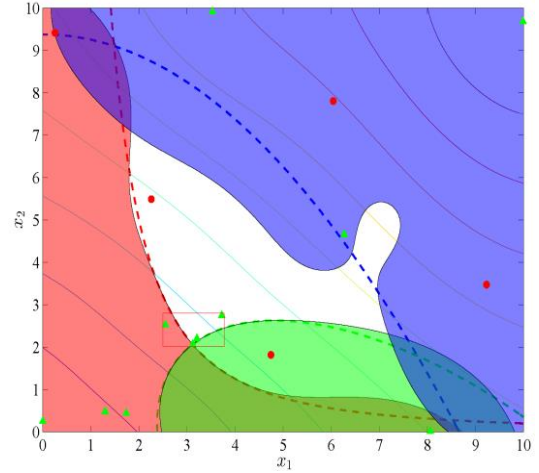

(f) iteration #9~12

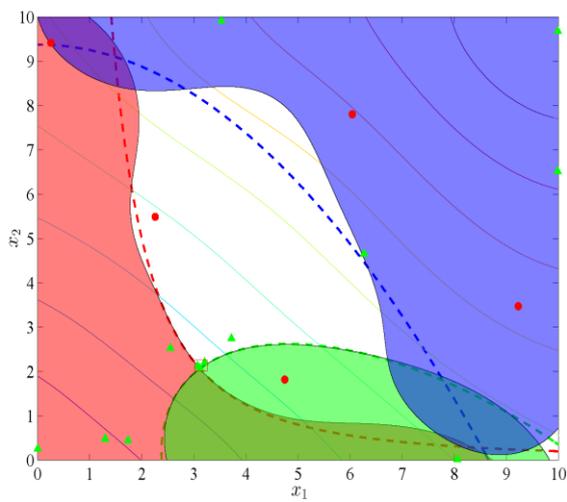

(g) iteration # 13-16

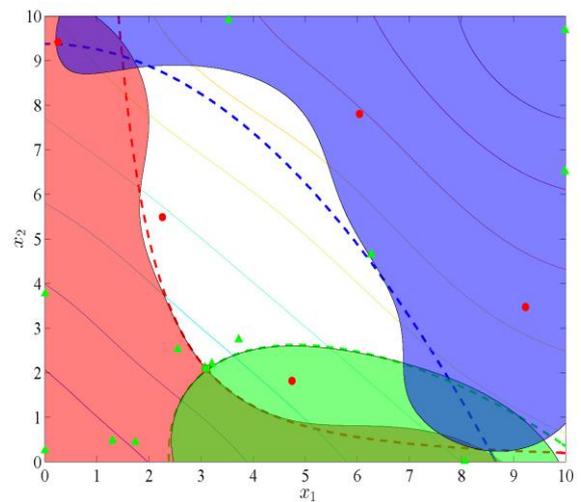

(h) iteration #17~18

Fig. 8 The iteration process of the surrogate models for the 2D problem. Where the dashed lines --, -- and -- are the contour lines determined by the real constraints $g_1(x) = 0$, $g_2(x) = 0$ and $g_3(x) = 0$; ▬, ▬, ▬ are the regions determined by the surrogate constraints $\hat{g}_1(x) \geq 0$, $\hat{g}_2(x) \geq 0$ and $\hat{g}_3(x) \geq 0$; the symbol ● denotes the initial samples, ▲ denotes the added samples; the curves ⦀ are the contour lines of the surrogate objective function $\hat{J}(x)$; ☐ is the reduced interval.

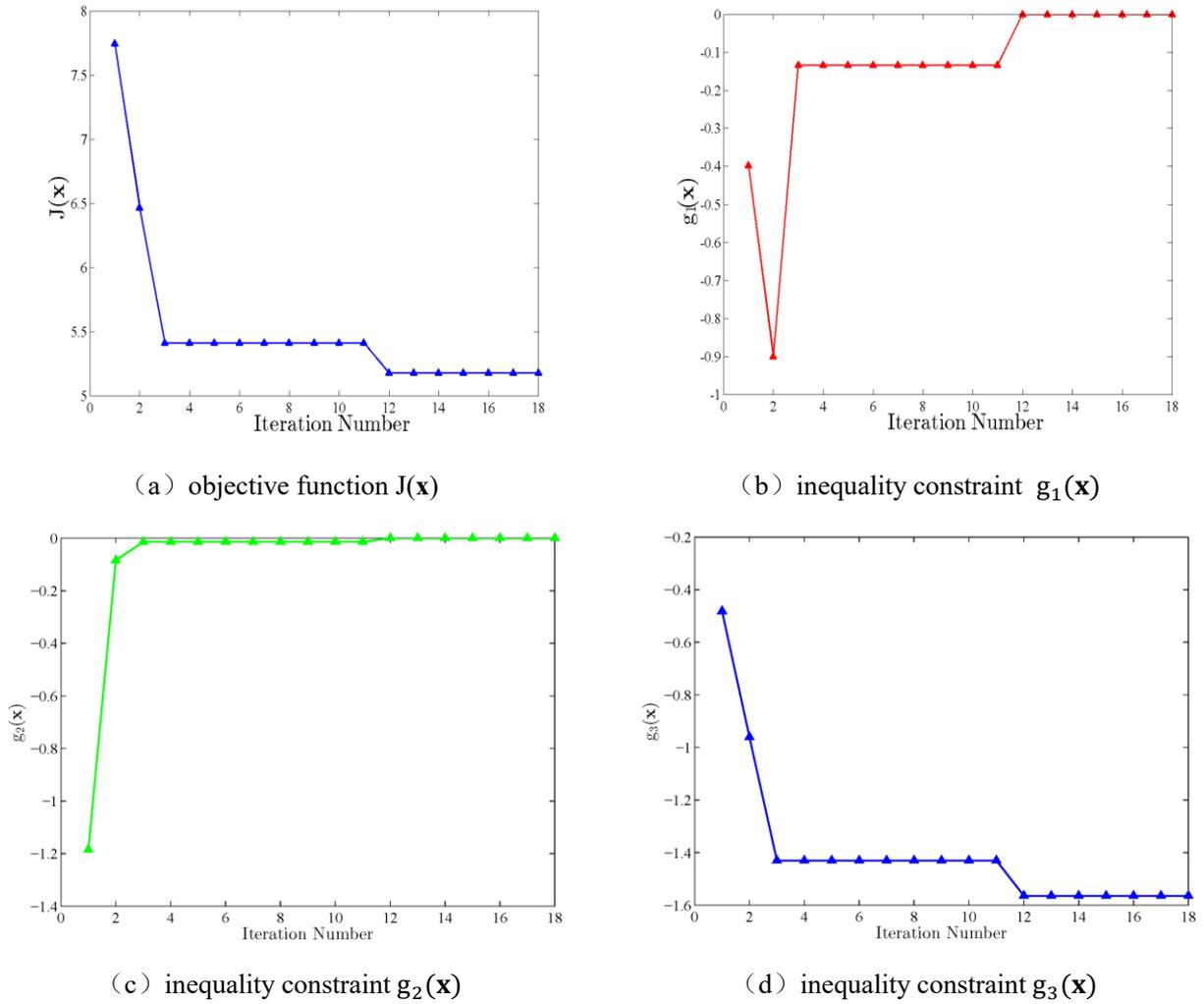

（a）objective function J(x)  　　　　　（b）inequality constraint $g_1(x)$

（c）inequality constraint $g_2(x)$  　　　　　（d）inequality constraint $g_3(x)$

Fig. 9 The iteration of the current objective and constraint functions for the 2D problem.

Table 3 Results of different methods for the 2D problem. (Average of 50 simulations)

| Methods | Mean value of J(x) | Standard deviation of J(x) | Number of function evaluations |
| --- | --- | --- | --- |
| SQP | 5.2580 | 0.5760 | 23 |
| GA | 5.1927 | 0.03486 | 3152 |
| SSBO | 5.1768 | 0.00572 | 17 |

Fig. 8 shows the iterative process of the surrogate models. It can be seen that at the beginning of the iteration, the errors of the surrogate models are large due to fewer sample points. As the iteration goes on, the sample points within the region including the global optimum are gradually increased, and the convergence is accelerated to the actual optimum through the interval reduction method. The active constraints $g_1(\mathbf{x})$ and $g_2(\mathbf{x})$ have high precision in the vicinity of the optimum because of a large number of sample points. Meanwhile, the non-active constraint function $g_3(\mathbf{x})$ has poor accuracy at the boundary line $g_3(\mathbf{x})=0$ due to the sparse sample points. Since the non-active constraint is far from the optimum, the poor accuracy will not affect the optimization, so the algorithm can effectively reduce the sample evaluations of the non-optimal region and improve the search efficiency.

Fig. 9 shows the iterations of the objective and constraint functions. It can be seen that the objective function at the beginning of iteration decreases rapidly and converges after about 12 iterations. When the active constraint functions $g_1(\mathbf{x})$ and $g_2(\mathbf{x})$ converge, they reach the 0 boundaries.

Table 3 shows the optimization statistics results of different algorithms. Since the problem has only one local optimum, all algorithms can converge to the global optimum. However, due to the different initial samples, the required function evaluations are quite different. In this problem, SSBO uses 17 function evaluations (including five initial samples) to converge to the optimum, and the convergence variance is small, which verifies the effectiveness and accuracy of the proposed method.

4.3 Welded plate structure optimization problem

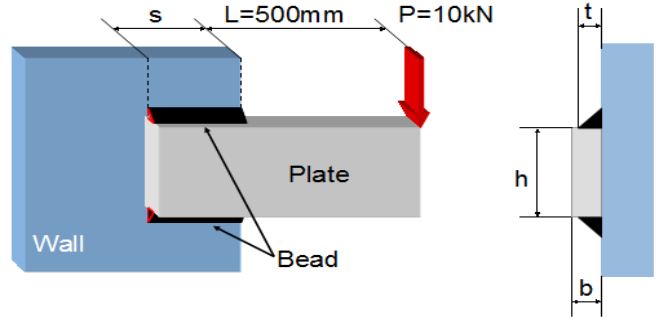

Fig. 10 Schematic diagram of the welded plate structure.

As shown in Fig. 10, the structural optimization problem of the welded plate includes four design variables and five constraint functions. By optimizing the geometric parameters of the welded plate, the cost function is minimized and the bending stress, shear stress, impact strength and geometric related constraints are satisfied. The optimization problem is defined as follows:

$$\begin{aligned} &\text{find } \mathbf{x} = [s,t,h,b]^T \\ &\min_{\mathbf{x}} \text{Cost}(\mathbf{x}) \\ &\text{s.t.} \begin{cases} g_1(\mathbf{x}) = \sigma_{\max}(\mathbf{x}) - 210\text{MPa} \leq 0 \\ g_2(\mathbf{x}) = \tau_{\max}(\mathbf{x}) - 70\text{MPa} \leq 0 \\ g_3(\mathbf{x}) = \delta_{\max}(\mathbf{x}) - 5.0\text{mm} \leq 0 \\ g_4(\mathbf{x}) = \eta_{P_c}(\mathbf{x}) \leq 0 \\ g_5(\mathbf{x}) = \eta_{TB}(\mathbf{x}) \leq 0 \end{cases} \end{aligned} \quad (17)$$

$$\text{Cost}(\mathbf{x}) = C_1 s t^2 + C_2 (L+s) h b \tag{18}$$

$$\sigma_{\max}(\mathbf{x}) = 6 P L b^{-1} h^{-2} \tag{19}$$

$$\begin{aligned}
&M(\mathbf{x}) = P(L+0.5s) \\
&R(\mathbf{x}) = \sqrt{0.25 s^2 + 0.25(t+h)^2} \\
&J(\mathbf{x}) = 2\sqrt{2}\, ts \left[ s^2/12 + 0.25(t+h)^2 \right] \\
&t_1(\mathbf{x}) = P\left(\sqrt{2}\, ts\right)^{-1} \\
&t_2(\mathbf{x}) = M(\mathbf{x}) R(\mathbf{x}) J(\mathbf{x})^{-1} \\
&\tau_{\max}(\mathbf{x}) = \sqrt{t_1(\mathbf{x})^2 + t_2(\mathbf{x})^2 + t_1(\mathbf{x}) t_2(\mathbf{x}) s R(\mathbf{x})^{-1}}
\end{aligned} \tag{20}$$

$$\delta_{\max}(\mathbf{x}) = 4 P L^3 b E^{-1} h^{-3} \tag{21}$$

$$\begin{aligned}
&t_3(\mathbf{x}) = \sqrt{E G h^2 b^6 / 36} \\
&t_4(\mathbf{x}) = 1.0 - 0.25 h L^{-1} \sqrt{E G^{-1}} \\
&P_c(\mathbf{x}) = 4.013\, t_3(\mathbf{x}) t_4(\mathbf{x}) L^{-2} \\
&\eta_{P_c}(\mathbf{x}) = 1.0 - P_c(\mathbf{x}) P^{-1}
\end{aligned} \tag{22}$$

$$\eta_{TB}(\mathbf{x}) = t b^{-1} - 1.0 \tag{23}$$

where $\mathbf{x}$ is the vector of design variables, $s \in [100, 500]$mm is the length of the welding point, $t \in [2.0, 6.0]$mm is the thickness of the welding point, $h \in [100, 500]$mm is the height of the welding point, $b \in [5, 10]$mm is the thickness of the welding plate. Cost($\mathbf{x}$) is the cost function. $C_1 = 6.739 \times 10^{-5}$ and $C_2 = 2.936 \times 10^{-6}$ are the weld cost coefficients. $L = 500$mm is the distance between load and wall. $g_i(\mathbf{x})(i = 1, 2, \ldots, 5)$ is the $i$th constraint function. $\sigma_{\max}(\mathbf{x})$(MPa) is the maximum bending stress. $\tau_{\max}(\mathbf{x})$(MPa) is the maximum shear stress. $\delta_{\max}(\mathbf{x})$(mm) is the maximum displacement. $\eta_{P_c}(\mathbf{x})$ is the impact strength ratio. $\eta_{TB}(\mathbf{x})$ is the geometric constraint ratio. $P=10000.0$N is the load. $G=82680$MPa is the shear modulus. Moreover, $M(\mathbf{x})$, $R(\mathbf{x})$, $J(\mathbf{x})$, $t_1(\mathbf{x})$, $t_2(\mathbf{x})$ and $P_c(\mathbf{x})$ are the intermediate variables defined for convenience of description.

The SQP and GA in the solution process are implemented by the optimization toolkit of Matlab2015a. In this example, the maximum number of optimizations for SSBO is set as $k_{\max}$=100, and the number of the initial sample points is $n_0$=9, which are sampled Latin hypercube sampling method. The calculation results are shown in Fig. 11.

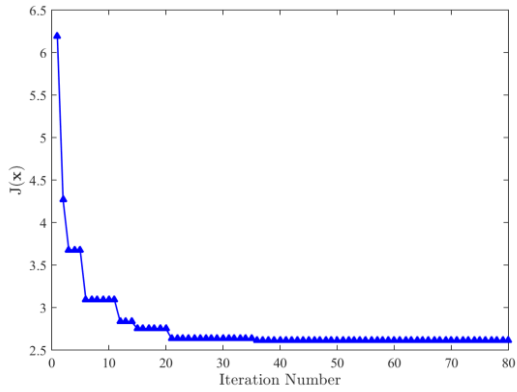

(a) The objective function Cost(**x**).

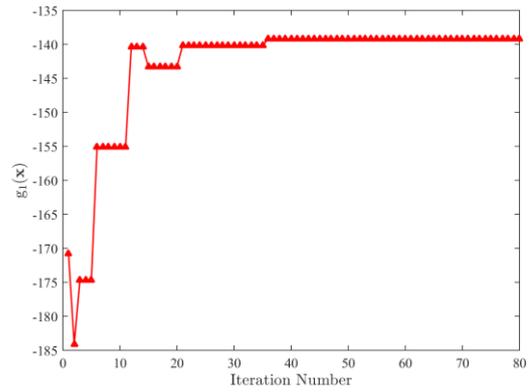

(b) The maximum bending stress constraint $g_1(\mathbf{x})$.

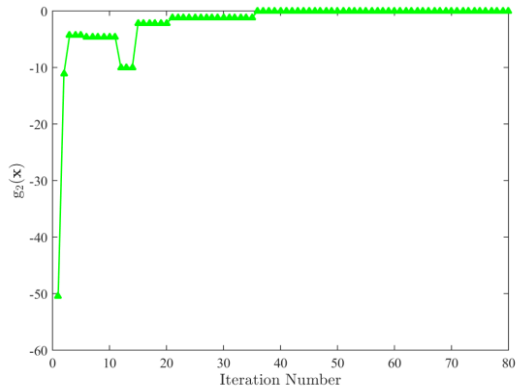

(c) The maximum shear stress constraint $g_2(\mathbf{x})$

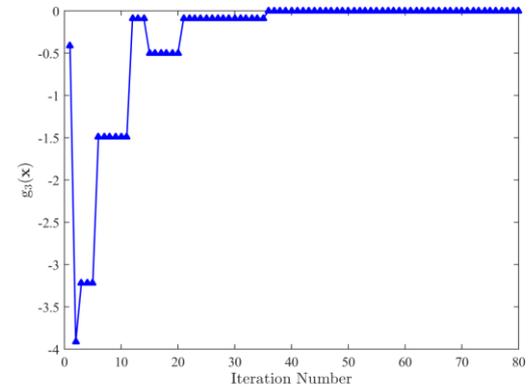

(d) The maximum displacement constraint $g_3(\mathbf{x})$.

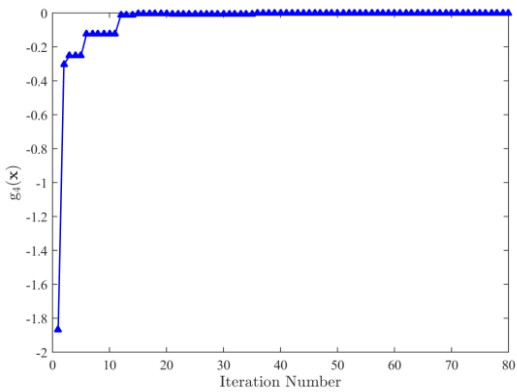

(e) The impact strength ratio constraint $g_4(\mathbf{x})$.

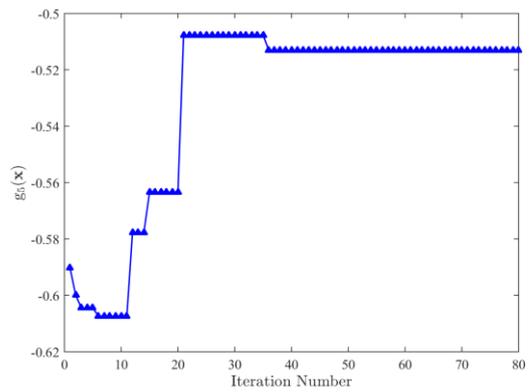

(f) The geometric ratio constraints $g_5(\mathbf{x})$.

Fig. 11 Iterations for the welded plate structure optimization problem.

Table 4 Results for the welded plate structure optimization problem.

| Methods | Mean value of Cost(**x**) | Standard deviation of Cost(**x**) | Number of function evaluations |
| --- | --- | --- | --- |
| SQP | 2.6143 | $2.3614 \times 10^{-7}$ | 215 |
| GA | 5.3000 | 0.5254 | 48191 |
| SSBO | 2.6332 | 0.03726 | 54 |

Fig. 11 gives the iterative process of the objective and constraint functions for the welding plate problem. As expected, the objective function Cost(**x**) gradually decreases with the iterative process, and the reduction process remains for a while and then suddenly becomes smaller, because not every added sample point improves the current optimum. It only improves when the samples reach a certain number. Meanwhile, it can be seen in Fig. 11 (c), (d) and (e) that as the iterative process goes on, the maximum shear force constraint $g_2(\mathbf{x})$, the maximum displacement constraint $g_3(\mathbf{x})$ and the impact strength ratio constraint $g_4(\mathbf{x})$ reach the critical value and therefore are active constraints. It can be seen in Fig. 11 (b), (f) that the maximum bending stress constraint $g_1(\mathbf{x})$ and the geometric constraint $g_5(\mathbf{x})$ are far from the constraint boundary value 0 and therefore are non-active constraints. In this example, GA is evaluated many times because of a large number of constraints, but the convergence effect is still poor. Although the mean and standard deviation of SSBO are not the smallest among several optimization methods, the number of function evaluations are only 0.11%~25.12% of the existing methods, and have higher convergence efficiency and convergence accuracy.

4.4 Wing structure optimization problem

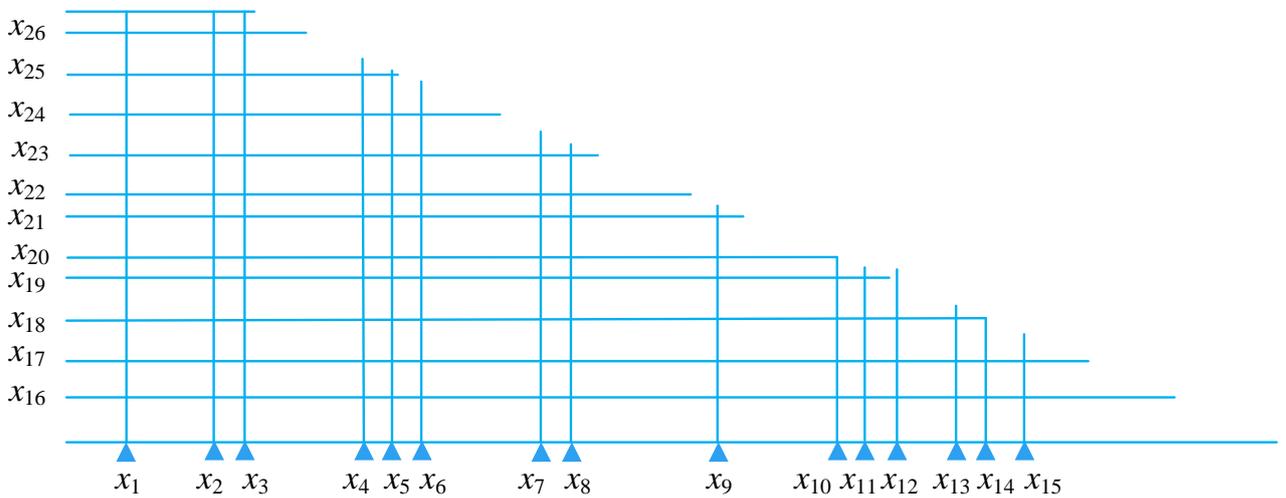

Fig. 12 The schematic diagram of the position of the wing beam and rib.

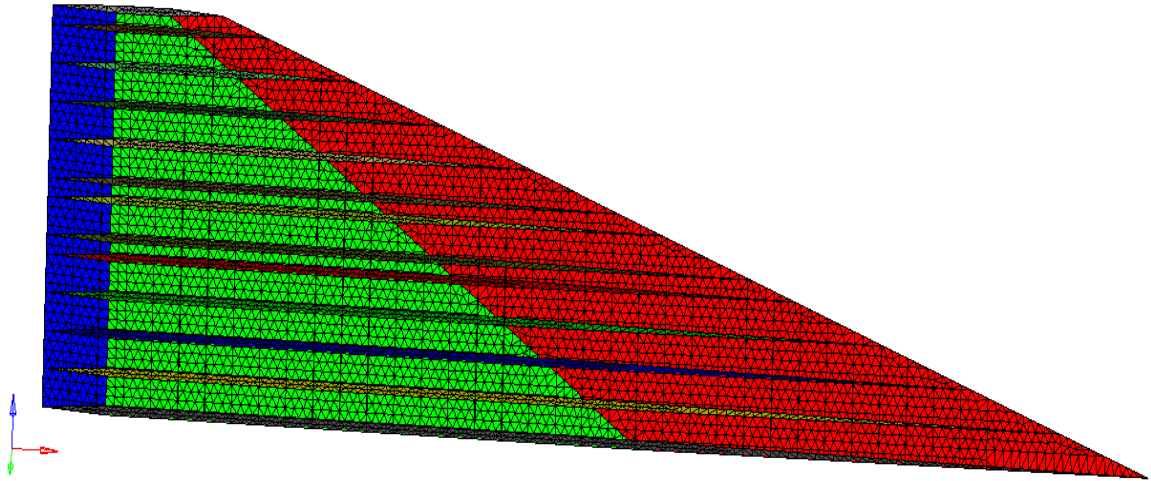

Fig. 13 The finite element mesh of the wing structure.

This example is the optimization of the wing structure of a high-speed aircraft: the design variables are the position of the wing structure beam and rib (as shown in Fig. 12); the objective function is to minimize the structural mass; the design constraint is that maximum stress in each part does not exceed the corresponding allowable stress. The structural model is a finite element analysis model as shown in Fig. 13. The model uses an unstructured grid with a total of 18,504 cells. The solver uses Nastran, and the average time of a single analysis is about 64.8s (Operating environment: Windows, 64 bits, 2GHz). The optimization problem is described as follows:

$$\min J(\mathbf{x}) = \text{Weight}(\mathbf{x})$$
$$\text{s.t.} \begin{cases} \mathbf{g}(\mathbf{x}) \leq \mathbf{0}, \mathbf{g}(\cdot) \in \mathbb{R}^{29} \\ \mathbf{x}_L \leq \mathbf{x} \leq \mathbf{x}_U, \mathbf{x} \in \mathbb{R}^{26} \end{cases} \quad (24)$$

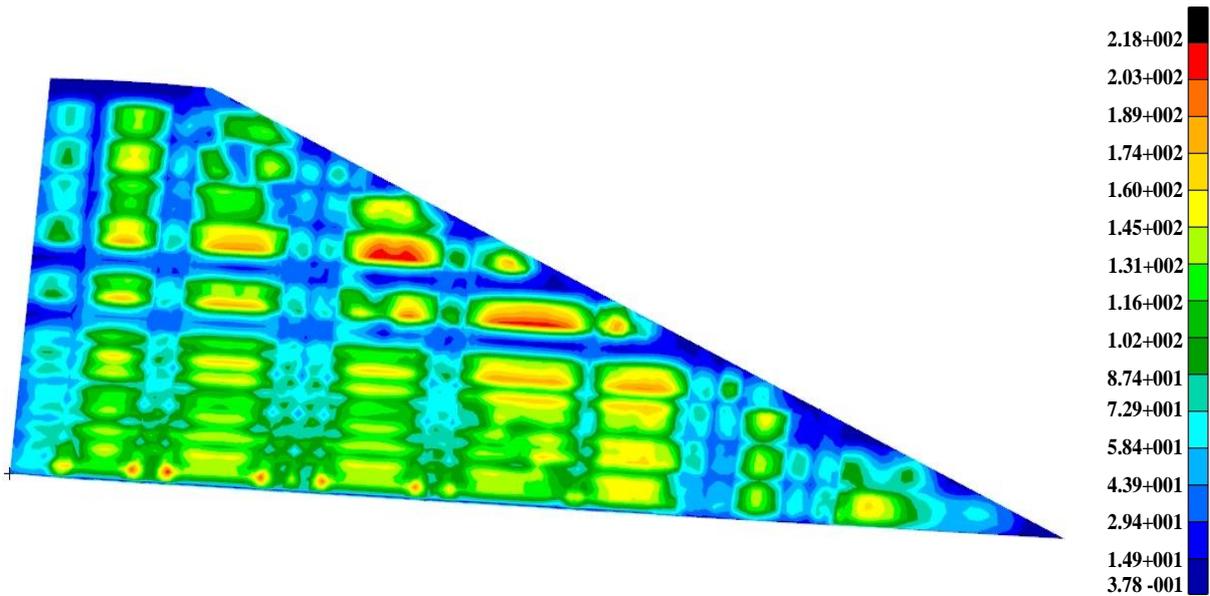

Fig. 14 The stress distribution of the optimized wing structure.

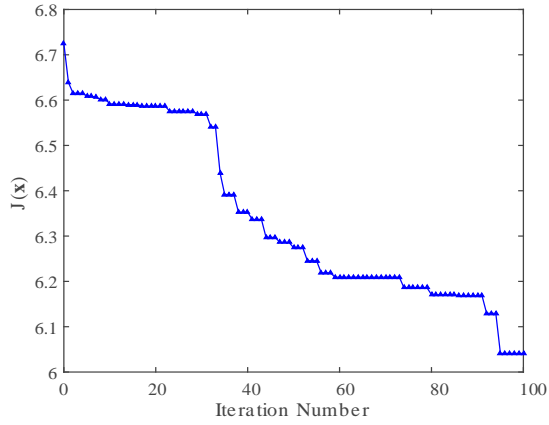
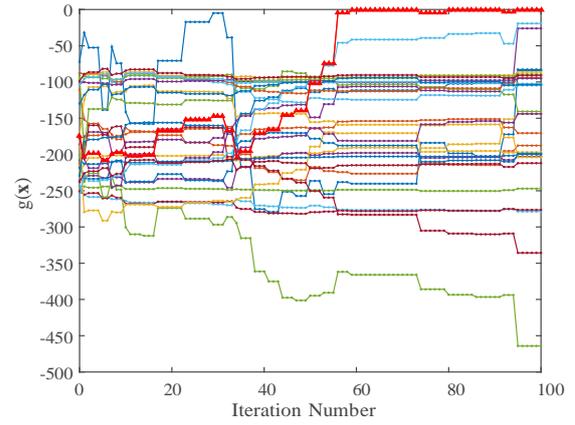

(a) Iteration of the mass of the wing structure.

(b) Iteration of the constraint function, where ─▲─ is the active constraint $g_{16}(x)$ and the other curves are the remaining constraint functions.

Fig. 15 Iterations of the objective and constraint functions for the wing structure optimization problem.

Table 5 Results for the wing structure optimization problem.

| Methods | Objective function | Number of function evaluations | Time consuming (h) |
| --- | --- | --- | --- |
| SQP | 6.6265 | 3020 | 54.37 |
| GA | 6.5947 | 50000 | 901.24 |
| SSBO | 6.0411 | 153 | 10.36 |

The stress distribution of the optimized wing structure is shown in Fig. 14. As shown in Fig. 15, the objective function value J(x) gradually decreases, and the active constraint function $g_{16}(x)$ gradually converges to the constraint boundary. Since the number of the initial sample points is 53, the SSBO has performed 153 function evaluations in total. As expected, the convergence process is not a continuous change, but a step-like change. The function value remains unchanged for a period of time until the number of sample points is increased, and the function value is abruptly changed to a certain extent. The added sample points need to be accumulated to a certain extent so that the local accuracy of the model is high enough. After the algorithm finds a better solution, the objective function value is improved. How many sample points need to be added to improve the process is related to the complexity of the original expensive black-box model, which is currently impossible to make a pre-estimation. As shown in Table 5, the objective function value obtained by the proposed method is significantly smaller than other methods in terms of optimal performance, so it has better global search ability than SQP and GA. From the perspective of computational efficiency, the number of function evaluations is small, and the overall calculation time of the algorithm is only 1.15%~19.05% of the comparison method, which illustrates the calculation efficiency has significant advantages. The calculation time of SSBO is mainly combined with three parts: first, the sample point evaluation is time-consuming, which is also the main time cost in the problem; second, the parameter optimization time of the model construction; third, the optimization process is time-consuming, Since the problem contains 26 design variables and 29 constraints, the calculation process takes more time when the optimization algorithm is used to increase the sample points, but the overall time still has significant advantages over the existing methods. Therefore, the proposed method has a strong global optimization ability in practical problems, and has high computational efficiency in solving expensive black-box optimization problems.

## 5 Conclusions

In this paper, a hybrid criterion to infill new samples asynchronously based on radial basis function (RBF) and interval reduction is proposed. In the proposed criterion, three sub-criteria are involved. First, searching for the global optimum with genetic algorithm (GA) based the current surrogate models of objective and constraint functions. Second, to exploit the sample space with a strict minimum distance constant, a grid searching with Latin hypercube sampling (LHS) with the current surrogate model is adopted. This sub-criterion aims to infill samples in the region with sparse samples. Third, searching for local optimum with sequential quadratic programming (SQP) based on the current surrogate model. Here, the searching space is a neighborhood of the current optimum in the existing sample set. The neighborhood is a promising region including a constant number of samples. The results of the examples show that SSBO has better global search capability and local convergence accuracy than the existing methods in the multi-peak problem. In the optimization problems with only one local optimum, there are fewer evaluations than existing methods. Therefore, SSBO has good global convergence ability, local convergence accuracy and computational efficiency. This method is not a general optimization algorithm, while the existing general optimization algorithms are its foundations. SSBO is only applicable to the expensive black-box optimization problem. In addition, subjective to the predictive ability of the surrogate model, SSBO can only deal with continuous optimization problems. In the future work, the mixed optimization problem with discrete variables can be considered, and how to establish an accurate surrogate model and how to add points in that case are interesting topics.

## Acknowledgment

This study is supported by the China Civil Aerospace Program (No. D010403), the National Defense Fundamental Research Funds of China (No. JCKY2016204B102).

283